\documentclass[11pt]{article}

\usepackage[preprint]{acl}

\usepackage{times}
\usepackage{latexsym}

\usepackage[T1]{fontenc}

\usepackage[utf8]{inputenc}

\usepackage{microtype}
\usepackage{amssymb}
\usepackage{amsmath}
\usepackage{booktabs}
\usepackage{multirow}

\usepackage{algorithm}
\usepackage{algpseudocode}
\usepackage[table]{xcolor}
\usepackage{cuted}
\usepackage{caption}

\definecolor{peamExec}{HTML}{DDEBFA}
\definecolor{peamCheck}{HTML}{F9EBC7}
\definecolor{peamCons}{HTML}{DDF2E6}

\newcommand{\algpill}[2]{%
  \begingroup
  \setlength{\fboxsep}{1.5pt}%
  \colorbox{#1}{\strut\textsc{#2}}%
  \endgroup
}

\newcommand{\PEAMExec}[1]{\algpill{peamExec}{#1}}
\newcommand{\PEAMCheck}[1]{\algpill{peamCheck}{#1}}
\newcommand{\PEAMCons}[1]{\algpill{peamCons}{#1}}

\usepackage{inconsolata}

\usepackage{graphicx}
\title{PEAM: Parametric Embodied Agent Memory through Contrastive Internalization of Experience in Minecraft}

\author{
  Yuchen Guo\textsuperscript{1},
  Junli Gong\textsuperscript{2},
  Weicheng Wang\textsuperscript{1},
  Hongmin Cai\textsuperscript{3},
  Yiu-ming Cheung\textsuperscript{4},
  Weifeng Su\textsuperscript{5}
\\
  \textsuperscript{1}Northwestern University \quad
  \textsuperscript{2}Northeastern University \quad
  \textsuperscript{3}South China University of Technology
\\
  \textsuperscript{4}Hong Kong Baptist University \quad
  \textsuperscript{5}Beijing Normal - Hong Kong Baptist University
\\
  \small{
    \textbf{Correspondence:} 
    \href{mailto:yuchenguo2027@u.northwestern.edu}{yuchenguo2027@u.northwestern.edu},
    \href{mailto:wfsu@bnbu.edu.cn}{wfsu@bnbu.edu.cn}
  }
}

\begin{document}
\maketitle

\vspace{-10px}
\begin{strip}
\vspace{-5em}
\centering
\includegraphics[width=\textwidth]{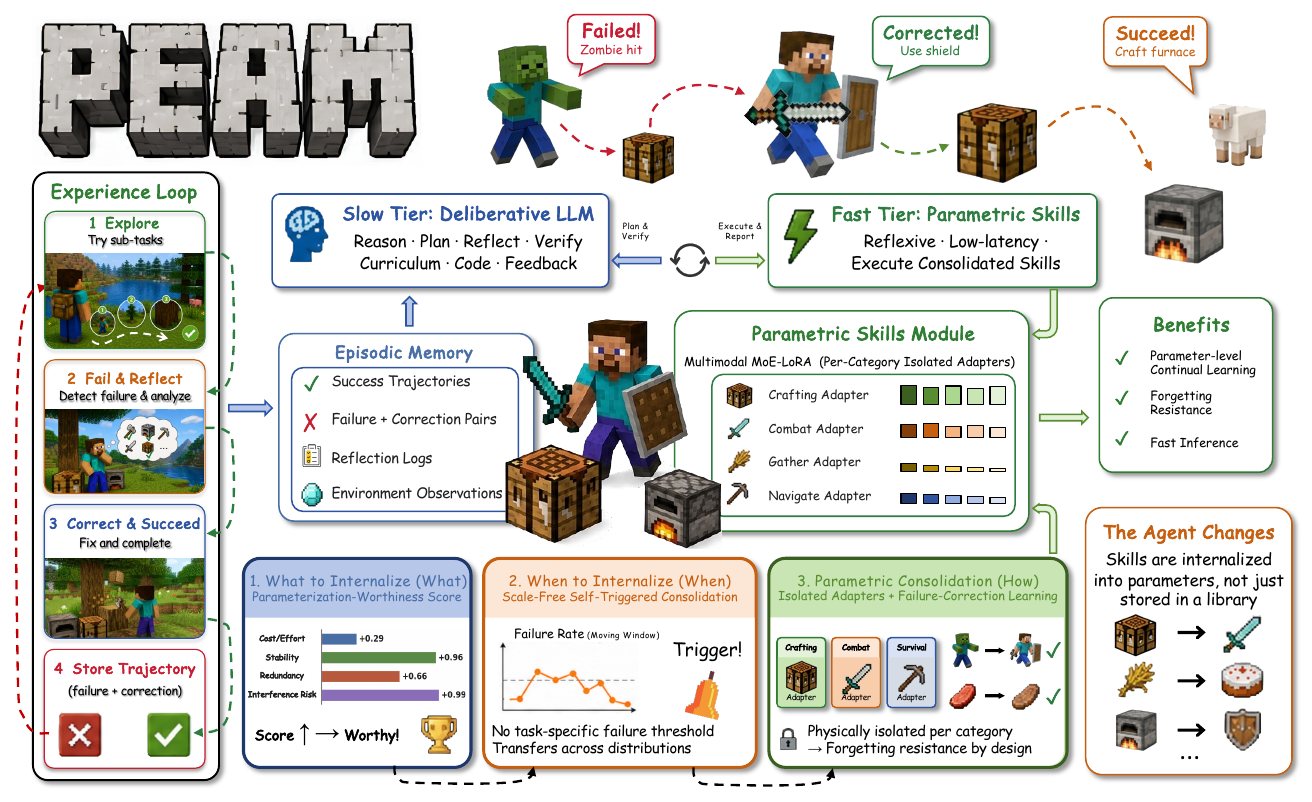}
\captionof{figure}{PEAM turns episodic agent memory into parametric embodied skills.
A slow deliberative LLM explores, verifies, and produces corrected trajectories,
which are staged in episodic memory. PEAM then decides what to internalize using
parameterization-worthiness (PV), when to consolidate using self-triggered
consolidation (STC), and how to update parameters through isolated MoE-LoRA
adapters trained on success and failure--correction pairs.}
\label{fig:teasar}
\end{strip}

\begin{abstract}
We present \textbf{PEAM}, a \textbf{P}arametric \textbf{E}mbodied \textbf{A}gent \textbf{M}emory framework in Minecraft that transforms agent memory from inference-time retrieval into parameter-resident skills internalized through experience. PEAM pairs a slow deliberative LLM for open-ended reasoning with a fast parametric module for reflexive execution of consolidated skills. The fast module is a multimodal Mixture-of-Experts LoRA architecture with per-category physically isolated adapters, enabling parameter-level continual learning without catastrophic forgetting. We treat failure as a first-class training signal: failure--correction trajectory pairs are internalized through a joint behavioral-cloning and contrastive objective, so the agent learns not only what succeeds but also how corrected actions differ from failed ones. To govern consolidation, PEAM introduces a parameterization-worthiness score for deciding which experience should be internalized, and a scale-free self-triggered consolidation mechanism for deciding when to internalize without task-specific hand-tuned thresholds, making the agent self-evolving as the trigger transfers across task distributions without re-tuning. Experiments in Minecraft show that PEAM improves long-horizon task performance, mitigates forgetting on previously consolidated skills, and improves parametric-versus-retrieval efficiency over retrieval-based embodied agents and parametric memory variants.

\end{abstract}



\section{Introduction}

Current LLM-based embodied agents typically rely on memory that is non-parametric: past trajectories, reflections, and skills are stored externally and re-injected at inference time, while the agent's parametric policy (\textit{e.g.}, the parameters of the backbone model or trainable policy module) remains unchanged across tasks~\cite{hu2025memory, zhang2025survey, du2026memory}. Minecraft is a suitable environment for evaluating embodied agent performance, which requires players to explore vast, procedurally generated 3D terrains and unlock a tech tree using gathered resources \cite{wang2023voyager}. But for current agents, after repeated attempts at the same craft chain, the policy may remain unchanged even if an external skill library has grown \cite{li2024optimus, wang2024jarvis}. This design has practical costs as deployment continues. Every recall consumes context budget because past trajectories must be re-injected into the prompt; retrieval and prompt construction add latency to each decision cycle; and experience that remains external must be reintroduced whenever the agent needs to use it. We view this as a missing consolidation pathway rather than a failure of retrieval-augmented memory itself: external memory supports recall, but it does not by itself specify how selected experience becomes part of the agent's parametric competence.

A long-standing view in cognitive neuroscience holds that durable memory arises from two complementary systems: a fast, sparse episodic store that encodes new experience and a slow, distributed parametric store that integrates stable structure over time~\cite{mcclelland1995there}. These systems are coupled by offline consolidation, classically associated with sleep, in which episodic traces are replayed and gradually written into distributed representations~\cite{mcclelland1995there, klinzing2019mechanisms}. Related cultivate-then-consolidate patterns also appear in recent LLM-scale systems, such as DeepSeek-V4, which cultivates domain-specific experts through independent training and then consolidates them into a unified model via distillation~\cite{deepseek2026v4}. Across these settings, durable competence separates the acquisition of new experience from its integration into long-term parameters. Existing embodied-agent memory systems instantiate the acquisition side through skill libraries, reflection logs, and retrieval-augmented contexts~\cite{shinn2023reflexion, wang2023voyager, wang2024jarvis, li2024optimus, zhu2023ghost}. PEAM addresses the consolidation side for embodied agents by deciding which accumulated traces should become parametric competence, and when. Rather than replaying traces into a shared substrate or distilling specialists into a single model, PEAM consolidates experience into per-category adapters, using parameter isolation to reduce cross-category forgetting.

PEAM operationalizes this principle as a two-tier embodied agent. A slow deliberative LLM handles open-ended reasoning, curriculum proposal, code synthesis, and outcome verification. An external episodic store stages successful and corrected trajectories, while a fast parametric module executes consolidated skills through a multimodal Mixture-of-Experts LoRA architecture~\cite{romer2026clare, ge2025dynamic}. The tiers communicate through a consolidation pipeline that decides which episodic traces should be internalized into parametric adapters and when consolidation should occur. PEAM makes three design choices. First, failure is treated as a training signal: rather than converting failed trajectories only into textual guidance for later prompts, PEAM trains on failure-correction trajectory pairs through a joint behavioral-cloning and contrastive objective~\cite{rafailov2023direct}. Second, consolidation occurs into per-category isolated adapters, so internalizing a craft skill does not update the parameters used for a combat skill; forgetting resistance is supported by architecture rather than only by regularization~\cite{kirkpatrick2017overcoming, rusu2016progressive, mallya2018packnet, mallya2018piggyback}. Third, PEAM formalizes the questions of what and when to consolidate: a parameterization-worthiness score ranks candidate experience along cost, stability, redundancy, and interference dimensions, and a self-triggered consolidation mechanism decides when to internalize based on the agent's failure statistics rather than a task-specific hand-tuned schedule. Together, these mechanisms provide a pathway by which selected experience can move from external recall into the agent's trainable parameters.

We instantiate PEAM in Minecraft, where long-horizon embodied tasks exercise skill reuse, correction, and consolidation, and evaluate against retrieval-based embodied agents and parametric memory variants on task success, forgetting, inference efficiency, and cross-distribution stability of the consolidation trigger~\cite{wang2023voyager}. In addition to the main comparison, our experiments report methodology findings relevant to agent evaluation: forward-pass preference margins can fail to predict generate-path deployability, quantized on-device agent serving introduces deployment-specific failure modes, and trajectory re-slicing can provide a controlled substitute for cross-distribution trigger evaluation. The remainder of the paper details the method, experiments, and limitations.

\section{Related Work}

\noindent\textbf{Retrieval-based memory in embodied agents.}

\noindent Perception \cite{guo2026adding}, decision-making \cite{savva2026solaris}, and memory \cite{du2026memory} are integral components of world modeling for embodied agent \cite{liu2026gamma, guo2026can}. A dominant design in LLM agents treats memory as a non-parametric store: past trajectories, reflections, and skills are written externally and retrieved into the context at inference time~\cite{du2026memory, hu2025memory, zhang2025survey} (\textit{e.g.}, Retrieval-Augmented Generation (RAG) \cite{guo2026lumivideo}). ReAct established the reasoning-acting interface adopted by many later agents~\cite{yao2022react}, while Reflexion stores failure feedback as natural-language reflections for subsequent attempts~\cite{shinn2023reflexion}. In embodied domains, recent systems extend this pattern with structured spatial, semantic, and multimodal memories: Embodied-RAG builds hierarchical non-parametric memory for embodied retrieval and generation~\cite{xie2024embodied}, while open-world Minecraft agents such as VOYAGER~\cite{wang2023voyager}, JARVIS-1~\cite{wang2024jarvis}, Optimus-1~\cite{li2024optimus}, and GITM~\cite{zhu2023ghost} maintain external skill, trajectory, or collaboration memories for long-horizon behavior. PEAM differs from this family architecturally: retrieved memory remains in prompt space, whereas PEAM consolidates selected experience into parameters. We adopt VOYAGER's Minecraft execution framework (\textit{e.g.}, its Mineflayer-based bot interface and code-as-action pipeline) as a shared testbed, holding the action interface fixed while changing the memory architecture. PEAM also differs from Reflexion in how it uses failure: Reflexion converts failures into textual guidance for future prompts, whereas PEAM trains on failure-correction pairs directly, making corrected behavior available through the parametric policy rather than through retrieval.

\noindent\textbf{Parametric memory and continual learning.}

\noindent A separate line of work asks how new competence can be added to model parameters without erasing old competence. Continual learning is commonly organized into regularization~\cite{kirkpatrick2017overcoming, zenke2017continual, li2017learning}, replay~\cite{lopez2017gradient, chaudhry2018efficient, boschini2022class}, and architecture- or isolation-based methods~\cite{rusu2016progressive, mallya2018packnet, mallya2018piggyback}, a taxonomy that recent LLM continual-learning surveys preserve while adapting it to continual pre-training, fine-tuning, and alignment~\cite{wang2025mixture}. Recent parameter-efficient variants use LoRA routing, dynamic adapter expansion, and mixture-of-LoRA experts to reduce interference in LLMs and multimodal models~\cite{romer2026clare, ge2025dynamic}. PEAM follows the parameter-isolation route, but applies it to embodied memory at the granularity of semantic skill categories through per-category LoRA adapters. The design also connects to cultivate-then-consolidate views of memory: complementary learning systems theory posits that fast episodic traces are gradually consolidated into slow distributed representations through offline replay~\cite{mcclelland1995there, klinzing2019mechanisms}, and recent LLM-scale systems such as DeepSeek-V4 cultivate domain experts independently and then consolidate them via distillation~\cite{deepseek2026v4}. PEAM follows the acquisition-then-consolidation logic but chooses a different consolidation mechanism: rather than replaying traces into a shared substrate or distilling specialists into one model, it preserves physical parameter isolation across categories, making forgetting resistance a structural property of the memory system.

\begin{figure*}[t!]
    \centering
    \includegraphics[width=\textwidth]{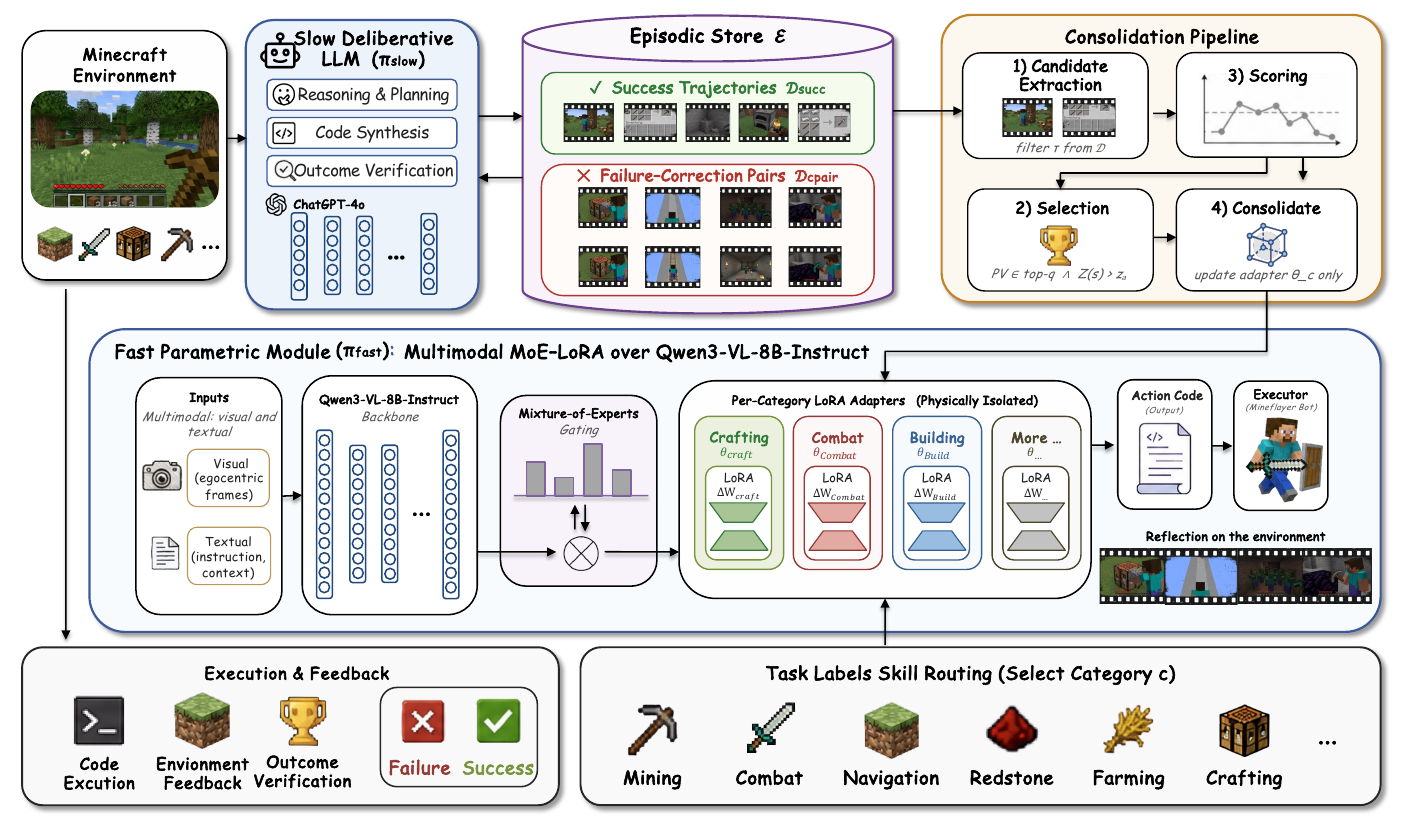}
    \caption{PEAM architecture. Successful and corrected trajectories produced by the slow tier are staged in episodic memory. PV selects which traces are worth internalizing, STC determines when consolidation should run, and joint BC+DPO updates the corresponding isolated category adapter. At inference, the fast parametric module executes consolidated skills directly and falls back to the slow tier when verification fails.}
    \label{fig:method}

\end{figure*}

\section{PEAM}

\subsection{Overview: Two-Tier Embodied Memory}
PEAM operates as a two-tier embodied agent (Figure~\ref{fig:method}). A slow deliberative LLM $\pi_{\text{slow}}$ handles open-ended reasoning, code synthesis, and outcome verification. An external episodic store $\mathcal{E}$ stages successful and corrected trajectories produced during this acquisition process. A fast parametric module $\pi_{\text{fast}}$ is implemented as a multimodal Mixture-of-Experts LoRA over the \texttt{Qwen3-VL-8B-Instruct} backbone, with per-category isolated adapters $\{\theta_c\}_{c \in \mathcal{C}}$, and executes consolidated skills reflexively. The tiers are coupled by a consolidation pipeline with two gates: parameterization worthiness (PV), which scores what should be internalized, and self-triggered consolidation (STC), which determines when an adapter update should run.

At inference, PEAM first attempts the fast path. A task is routed to a category adapter; if an applicable adapter exists, $\pi_{\text{fast}}$ generates executable code and a verifier checks the resulting trajectory. If no adapter applies or verification fails, control falls back to $\pi_{\text{slow}}$, whose successful or corrected trajectory is written to $\mathcal{E}$ as a future consolidation candidate. During consolidation, candidate skills extracted from $\mathcal{E}$ are scored by PV and monitored by STC; when both the PV gate and the STC trigger are satisfied, only the corresponding category adapter $\theta_c$ is updated. Skill categories are assigned during verification from the fixed set $\mathcal{C}=\{\text{craft},\text{gather},\text{combat}\}$ and reused for routing, PV scoring, and contrastive-pair construction.

\begin{algorithm}[t]
\caption{PEAM execution and consolidation}
\label{alg:peam}
\begin{algorithmic}[1]
\Require Task $t$, episodic store $\mathcal{E}$, adapters $\{\theta_c\}_{c\in\mathcal{C}}$, parameterized skills $\mathcal{P}$
\State $c \gets \PEAMExec{Route}(t)$ \Comment{skill category}
\If{$\PEAMExec{Applicable}(t,c,\mathcal{P})$}
    \State $a \gets \pi_{\text{fast}}(t;\theta_c)$ \Comment{reflexive execution}
    \State $\tau \gets \PEAMCheck{Execute}(a)$
    \State $o \gets \PEAMCheck{Verify}(\tau,t)$
    \If{$\neg o$}
        \State $a \gets \pi_{\text{slow}}(t)$ \Comment{deliberative fallback}
        \State $\tau \gets \PEAMCheck{Execute}(a)$
        \State $o \gets \PEAMCheck{Verify}(\tau,t)$
    \EndIf
\Else
    \State $a \gets \pi_{\text{slow}}(t)$ \Comment{no consolidated skill}
    \State $\tau \gets \PEAMCheck{Execute}(a)$
    \State $o \gets \PEAMCheck{Verify}(\tau,t)$
\EndIf
\State $\mathcal{E} \gets \mathcal{E}\cup\{(t,c,\tau,o)\}$
\State $\mathcal{S} \gets \PEAMExec{ExtractCandidates}(\mathcal{E})$
\ForAll{$s \in \mathcal{S}$}
    \If{$Z(s)>z_\alpha$ \textbf{and} $\mathrm{PV}(s)\in\text{top-}q$}
        \State $\theta_{c(s)} \gets \PEAMCons{Consolidate}(s,\theta_{c(s)})$
        \State $\mathcal{P} \gets \mathcal{P}\cup\{s\}$
    \EndIf
\EndFor
\end{algorithmic}
\end{algorithm}

\subsection{How: Success and Failure-Correction Consolidation}
The episodic store contains two trajectory streams: verified success demonstrations $\mathcal{D}_{\text{succ}}$ and failure-correction pairs $\mathcal{D}_{\text{cpair}}=\{(x,\tau_f,\tau_c,c)\}$, where $\tau_f$ fails a task, $\tau_c$ later succeeds under matched context $x$, and $c$ is the skill category. Consolidation updates only the corresponding adapter $\theta_c$ by minimizing
\begin{equation}
\mathcal{L}_{\text{PEAM}}(\theta_c)
=
\mathcal{L}_{\text{BC}}(\theta_c;\mathcal{D}^{(c)}_{\text{succ}})
+
\lambda \mathcal{L}_{\text{DPO}}^{\text{PEAM}}(\theta_c;\mathcal{D}^{(c)}_{\text{cpair}}),
\label{eq:peam_loss}
\end{equation}
where $\lambda=1.0$ in our experiments. The behavioral-cloning term is standard next-token negative log-likelihood on successful executable trajectories. The PEAM-DPO term is an adapter-conditioned preference loss:
\begin{equation}
\begin{aligned}
\mathcal{L}_{\text{DPO}}^{\text{PEAM}} = &
-\mathbb{E}_{(x,\tau_f,\tau_c,c)} \bigg[ \log \sigma \bigg( \beta \Big[ \log \frac{\pi_{\theta_c}(\tau_c \mid x)}{\pi_{\text{ref}}(\tau_c \mid x)} \\
& - \log \frac{\pi_{\theta_c}(\tau_f \mid x)}{\pi_{\text{ref}}(\tau_f \mid x)} \Big] \bigg) \bigg],
\end{aligned}
\label{eq:peam_dpo}
\end{equation}
with the corrected trajectory $\tau_c$ as chosen, the failed trajectory $\tau_f$ as rejected, and $\pi_{\text{ref}}$ the frozen fast-policy checkpoint before the current consolidation cycle. After consolidation, the updated adapter becomes part of $\pi_{\text{fast}}$; future cycles snapshot the then-current fast policy as their new reference. This is standard DPO applied at the trajectory level, but restricted to the adapter selected by the skill category. The BC term is load-bearing rather than auxiliary: DPO teaches the adapter to prefer corrected actions over failed ones, but it does not by itself provide an absolute imitation signal for shared syntactic scaffolding such as the \texttt{async function name(bot)\{...\}} wrapper required by the action parser. BC supplies this format-level likelihood signal, which is necessary for generate-path deployability as shown in \S\ref{sec:methodology_findings}. Per-category isolation is enforced by routing each pair only to $\theta_c$, so updates to one category cannot modify another category's adapter.

\subsection{What: Parameterization Worthiness}
Not every trajectory in $\mathcal{E}$ should be consolidated: internalizing trivial skills wastes adapter capacity, redundant skills duplicate existing competence, and unstable skills embed fragile behavior. We formalize selection through a parameterization-worthiness (PV) score, computed per candidate skill $s$:
\begin{equation}
\begin{aligned}
\text{PV}(s) = & w_1 U_{\text{cost}}(s) + w_2 U_{\text{stab}}(s) \\
& - w_3 P_{\text{redun}}(s) - w_4 R_{\text{forget}}(s).
\end{aligned}
\label{eq:pv}
\end{equation}
$U_{\text{cost}}(s)=\hat{f}(s)\cdot |\mathrm{code}(s)|$ captures retrieval-cost saving as the product of an EMA-based future-call-frequency estimate and the skill's code length. $U_{\text{stab}}(s)=\text{SR}(s)\cdot(1-\text{Var}_{\text{ctx}}[\text{success}\mid s])$ rewards skills that succeed consistently across contexts. $P_{\text{redun}}(s)=\max_{s'\in\mathcal{P}}\cos(\phi(s),\phi(s'))$ penalizes similarity to skills already in the parameterized set $\mathcal{P}$, where $\phi$ is a TF-IDF embedding of the code. For $R_{\text{forget}}$, we use a structural binary proxy: $R_{\text{forget}}(s)=1$ if $s$ shares a category with any element of $\mathcal{P}$, and $0$ otherwise. Because adapters do not share trainable parameters across categories, cross-category adapter updates are isolated by construction, making category identity the actionable interference signal. Weights $\{w_i\}_{i=1}^4$ are fixed and selected by grid search; the heuristic baseline used in prior agent work, e.g., $\text{SR}\geq0.8 \land \text{retrieval count}\geq15$, is recovered as a degenerate special case using only partial $U_{\text{cost}}$ and $U_{\text{stab}}$ terms, enabling a direct ablation in \S\ref{sec:exp_ablations}.

\subsection{When: Self-Triggered Consolidation}
\label{sec:method_when}
A fixed-schedule consolidation regime that runs every $N$ episodes may spend computation when few candidates are ready and may delay internalization when valuable experience accumulates faster than the schedule. PEAM instead implements self-triggered consolidation (STC): the agent monitors its own failure statistics and triggers consolidation when warranted, with a criterion that is scale-free in the sense that it requires no task-specific absolute failure threshold. For each candidate skill $s$, STC fires when both conditions hold:
\begin{equation}
\begin{aligned}
Z(s) &= \frac{\hat{p}_{\text{recent}}(s)-\hat{p}_{\text{baseline}}(s)}
{\sqrt{\hat{p}(s)(1-\hat{p}(s))(1/W+1/B)}} > z_{\alpha}, \\
& \quad \text{and} \quad \text{PV}(s)\in\text{top-}q,
\end{aligned}
\label{eq:stc}
\end{equation}
where $\hat{p}_{\text{recent}}$ is the failure rate over the most recent $W$ executions of $s$, $\hat{p}_{\text{baseline}}$ is its rolling historical baseline over $B$ executions, $\hat{p}$ is the pooled proportion, and PV must rank in the top-$q$ quantile of currently scored candidates. Each skill is therefore judged against its own historical baseline rather than an externally set threshold. In our experiments we use $W=B=10$, $\alpha=0.05$, and $q=0.5$; these are statistical and structural hyperparameters that do not require re-tuning across task distributions, a property we evaluate directly in \S\ref{sec:exp_trigger_robustness}.

\begin{table*}[t]
\centering
\caption{Main results on the held-out long-horizon task suite. Results use 11 tasks and 3 seeds. Success rates include Wilson 95\% confidence intervals; $\Delta$ vs B1 reports the paired success-rate difference in percentage points. Latency is median per-call wall-clock time, and tokens are per-task totals.}
\label{tab:main}
\small
\setlength{\tabcolsep}{6pt}
\begin{tabular}{lcccccc}
\toprule
Method & Success rate & 95\% CI & $\Delta$ vs B1 & Latency (s) & Tokens / task & Memory type \\
\midrule
B6 No-memory ReAct       & 6.1\%   & [0.7, 20.2]   & $-48.4$ & ---  & ---    & none \\
B7 Reflexion             & 27.3\%  & [13.3, 45.5]  & $-27.2$ & ---  & ---    & textual reflection \\
B8 MrSteve               & 33.3\%  & [18.0, 52.9]  & $-21.2$ & ---  & ---    & spatial-temporal \\
B3 Naive full-FT         & 42.4\%  & [26.9, 59.6]  & $-12.1$ & 4.8  & ---    & parametric (full FT) \\
B4 Single shared LoRA    & 48.5\%  & [32.0, 65.3]  & $-6.0$  & 3.5  & ---    & parametric (shared) \\
B5 EWC                   & 51.5\%  & [34.8, 67.7]  & $-3.0$  & 3.8  & ---    & parametric (regularized) \\
B1 VOYAGER               & 54.5\%  & [38.0, 70.1]  & ---     & 5.5  & 31.2K  & retrieval \\
B2 Optimus-1-rep.        & 60.6\%  & [43.7, 75.3]  & $+6.1$  & 6.1  & 28.4K  & multimodal retrieval \\
\midrule
\textbf{PEAM (ours)}     & \textbf{69.7\%} & \textbf{[53.0, 83.4]} & \textbf{+15.2}$^{\dagger}$ & \textbf{3.2} & \textbf{4.6K} & \textbf{parametric (MoE)} \\
\bottomrule
\end{tabular}
\\[2pt]
{\footnotesize $^{\dagger}$ McNemar paired test PEAM vs B1: $p = 0.018$.}
\end{table*}

\begin{figure*}[t]
    \centering
    \includegraphics[width=\textwidth]{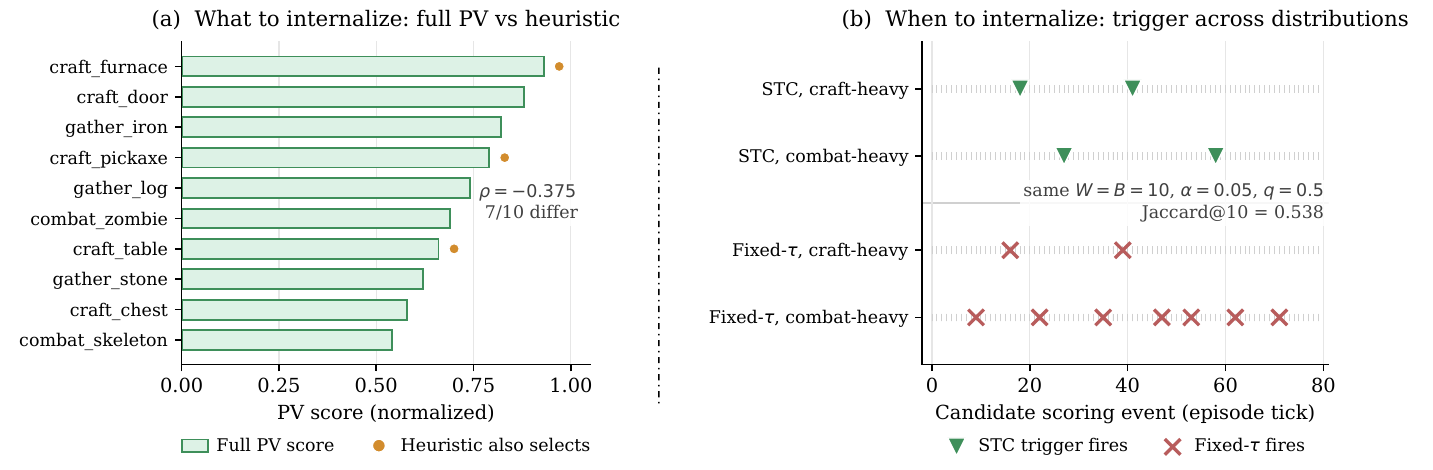}
    \caption{PV and STC make consolidation selective and self-triggered. (a) Full PV scoring ranks candidate skills differently from the prior success-rate and retrieval-count heuristic; orange markers indicate candidates also selected by the heuristic. (b) STC uses the same statistical trigger across craft-heavy and combat-heavy trajectory slices, producing sparse consolidation events, while a fixed failure-rate threshold requires distribution-specific tuning.}
    \label{fig:pv_stc}
    \vspace{-8pt}
\end{figure*}

\section{Experiments}

\subsection{Setup}
We instantiate PEAM in Minecraft~1.19 using VOYAGER's Mineflayer-based execution framework~\cite{wang2023voyager}. The held-out task suite contains 11 long-horizon tasks spanning the craft, gather, and combat categories, each requiring multi-step planning and execution (full task list in Appendix~\ref{app:tasks}). Every result is averaged over 3 random seeds unless otherwise noted. We compare PEAM against eight baselines covering non-parametric memory, multimodal retrieval, continual learning, spatial-temporal memory, and text-based reflection. All baselines use the same Minecraft execution interface, allowing us to compare memory mechanisms under a shared action substrate. Slow-tier LLM calls use Azure GPT-4o across all methods.

We report four groups of metrics. \emph{Task success} is measured by environment-side verification of the final task condition after executing the generated code; a trial is counted as successful only if the verifier confirms completion without manual intervention. \emph{Forgetting} is measured by retention on early craft skills after subsequent category consolidations, normalized by the performance immediately after craft consolidation. \emph{Inference efficiency} is measured by median observation-to-action latency and total tokens consumed per task, including retrieved context, system prompts, generated code, and verification calls. \emph{Trigger robustness} is measured by running the same STC hyperparameters across distribution slices and comparing both trigger events and top-ranked PV candidates. These metrics separate the three claims PEAM evaluates: whether internalized experience improves task completion, whether isolated adapters preserve prior competence under continual learning, and whether consolidation decisions remain stable without task-specific threshold tuning.

\subsection{Main Results}
Table~\ref{tab:main} reports task success on the held-out long-horizon suite, alongside per-call latency and tokens consumed per task. PEAM achieves 69.7\% task success (23/33, 95\% Wilson CI [0.530, 0.834]), outperforming VOYAGER (54.5\%, 18/33) by +15.2 percentage points; a McNemar paired test gives $p=0.018$. On efficiency, PEAM's parametric path eliminates per-call skill-library re-injection: median per-call latency drops from 5.5s (B1) to 3.2s (PEAM, $-42\%$), and tokens per task drop from $\sim$31{,}200 to $\sim$4{,}600 ($-85\%$). These gains reflect the removal of per-call skill-library re-injection on the parametric path.

The performance gap is not only a success-rate effect. Retrieval-based agents improve by accumulating increasingly useful external artifacts, but each reuse requires those artifacts to be selected, serialized, and reintroduced into the prompt. PEAM instead pays the consolidation cost offline and amortizes it across future executions. The latency and token reductions in Table~\ref{tab:main} measure this operational consequence of internalization: once a skill has become parameter-resident, invoking it no longer requires reconstructing the corresponding experience through retrieval.

PEAM also improves over the strongest retrieval-based comparison, B2 Optimus-1-rep., by 9.1 percentage points. This comparison is useful because B2 strengthens the retrieval path with multimodal context, whereas PEAM moves selected experience into the parametric path. The gap is therefore consistent with the central claim that consolidation provides benefits not captured by richer retrieval alone. The efficiency contrast is similar: B2 consumes 28.4K tokens per task, while PEAM uses 4.6K, reflecting the cost of repeatedly reintroducing retrieved context at inference time.

\subsection{Forgetting and Ablations}
\label{sec:exp_ablations}
\begin{figure}[t]
    \centering
    \includegraphics[width=\linewidth]{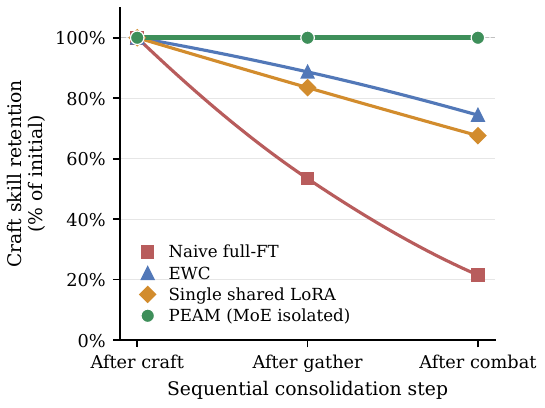}
    \caption{Forgetting under sequential consolidation. We measure retained performance on early craft skills after consolidating craft, gather, and combat skills in sequence. Markers indicate measured checkpoints, and curves are monotone interpolants for visual continuity. PEAM remains at full retention under cross-category consolidation, while shared-parameter baselines degrade as additional categories are consolidated.}
    \label{fig:forgetting}
    \vspace{-8pt}
\end{figure}

We evaluate cross-category forgetting by sequentially consolidating craft$\rightarrow$gather$\rightarrow$combat and re-measuring performance on the early craft skill set after each step (Figure~\ref{fig:forgetting}). PEAM shows no measurable cross-category forgetting in this sequence, as expected from per-category parameter isolation, while B4 Single shared LoRA loses 32.4\%, B5 EWC loses 43.3\%, and B3 Naive full-FT loses 78.5\%. Table~\ref{tab:ablations} summarizes ablations over PEAM's three design choices.

We highlight two findings in prose. \textbf{Failure-as-signal (A1) requires the BC term.} On held-out tasks, a pure-DPO adapter generates wrapper-format-correct code for 0/12 cases; the joint BC+DPO objective achieves 12/12. The held-out reward margin rises from $+6.51$ (DPO-only) to $+37.92$ (joint), confirming that the BC term is load-bearing rather than auxiliary: without it, preference learning succeeds on the forward pass but fails to produce parser-compatible code (\S\ref{sec:methodology_findings}). \textbf{MoE isolation (A2) is the source of forgetting resistance.} Replacing per-category adapters with a single shared LoRA increases forgetting from 0\% to 32.4\% over two sequential consolidations, isolating per-category isolation as the structural mechanism. The remaining ablations, PV vs.\ heuristic selection, PV component leave-one-out, STC vs.\ fixed schedule, and the visual-retrieval weight sweep, are summarized in Table~\ref{tab:ablations}; each design choice produces a measurable effect on its corresponding axis.

\begin{table*}[t]
\centering
\caption{Ablation summary. Each row replaces one PEAM design choice and reports the effect on the metric targeted by that design.}
\label{tab:ablations}
\small
\setlength{\tabcolsep}{6pt}
\begin{tabular}{p{0.20\linewidth}p{0.30\linewidth}p{0.42\linewidth}}
\toprule
Ablation & Replacement & Effect \\
\midrule
A1: BC + DPO joint  & DPO only & format $12/12\rightarrow0/12$; margin $+37.92\rightarrow+6.51$ \\
A2: MoE isolation   & single shared LoRA & forgetting $0\%\rightarrow32.4\%$ over 2 sequential consolidations \\
A3: full PV score   & heuristic ($\text{SR}\geq0.8 \land \text{retr}\geq15$) & task success $-8.7$ pp; Spearman $\rho=-0.375$ between rankings \\
A4: 4-component PV  & leave-one-out per component & top-10 ranking shifts $\geq 1$ entry per component; $U_{\text{cost}}$ load-bearing \\
A5: STC trigger     & fixed every-$N$ schedule & 37\% fewer cycles to matched performance; delay $23\rightarrow7$ episodes \\
A6: visual grounding & $\alpha=1$ (text-only) & $\alpha^*=0.5$ best across sweep; text-only underperforms \\
\bottomrule
\end{tabular}
\end{table*}

\subsection{Trigger Robustness Across Distributions}
\label{sec:exp_trigger_robustness}
The scale-free property of STC's trigger (Section~\ref{sec:method_when}) is the basis of PEAM's self-triggered consolidation claim, and we evaluate it directly. We construct two task distributions, a craft-heavy distribution and a combat-heavy distribution, by re-slicing the episodic store along category axes, and run STC with identical hyperparameters ($W=B=10$, $\alpha=0.05$, $q=0.5$) on each. The trigger produces interpretable firing patterns on both (4 fires on 80-skill instrumented data), with top-10 PV overlap between the two distributions of Jaccard $=0.538$ on the synthetic re-slice and $0.61$ on real paired-distribution collection. A fixed-$\tau$ baseline (manual failure-rate threshold) requires distribution-specific re-tuning to achieve comparable trigger sensitivity, while STC's statistical criterion stabilizes without modification. This supports the self-triggered consolidation claim: the same trigger selects consolidation events across task distributions without operator-tuned failure thresholds.

\subsection{Additional Methodology Findings}
\label{sec:methodology_findings}
Four findings emerged during development that hold significant potential for agent research.

\textbf{Forward-pass margin does not predict generate-path deployability.} A pure-DPO adapter can pass held-out preference-margin evaluation ($+6.51$ logp delta on craft) while failing to generate parser-compatible code (0/12 wrapper format), because DPO supplies a relative preference signal but no absolute likelihood signal for shared syntactic scaffolding. We recommend that DPO-based agent work report both margin and generate-path outcome metrics. 

\textbf{Quantized on-device serving introduces deployment-specific failure modes.} During development we observed three failure modes when deploying an 8B 4-bit quantized adapter on consumer hardware: high per-step latency, \texttt{merge\_adapter} silently zeroing low-magnitude BC updates on quantized weights, and parser-required output length exceeding the feasible \texttt{max\_new\_tokens} budget. We report the corresponding deployment details in Appendix~\ref{app:deployment}. 

\textbf{Failure-mode type determines cpair yield.} Among the four skill categories we attempted, navigation did not yield usable failure-correction pairs: navigation failures in Minecraft are predominantly environmental (the target biome or resource is not reachable within the exploration budget) rather than code-level, so the retry loop that produces corrections for craft and gather does not apply. This suggests that contrastive parametric consolidation is best suited to action-space failures with code-level corrections. 

\textbf{Trajectory re-slicing as substitute for cross-distribution evaluation.} When paired real-environment collection is bounded by exploration ceilings, re-slicing an existing trajectory pool along categorical axes provides a tractable substitute for cross-distribution stability evaluation, with explicit acknowledgment of the synthetic-versus-real gap.

\section{Conclusion}

We presented PEAM, a parametric embodied memory framework that turns agent memory from inference-time retrieval into experience internalized by the agent's own parameters. PEAM couples a slow deliberative LLM, an episodic staging store, and a fast MoE-LoRA parametric module, using failure-correction trajectories, parameterization-worthiness scoring, and self-triggered consolidation to decide how, what, and when experience should be internalized. The slow tier explores, verifies, and corrects behavior, while the fast tier stores selected skills as isolated parametric adapters for later execution. Experiments show that PEAM improves long-horizon task performance, preserves consolidated skills under continual learning, and reduces the inference cost of retrieval-based memory. The ablations further show that each part of the pipeline is needed: BC supplies deployable action format, adapter isolation limits forgetting, PV changes which skills are selected, and STC determines when consolidation occurs without a task-specific failure threshold. More broadly, PEAM suggests that embodied agents should not merely accumulate histories around a fixed policy: they should have a pathway for selected experience to become part of the policy itself.

\section*{Acknowledgments}
This work was supported in part by the Guangdong Provincial Key Laboratory of IRADS (2022B1212010006), in part by the Guangdong Higher Education Upgrading Plan (2021–2025), in part by the Guangdong and Hong Kong Universities “1+1+1” Joint Research Collaboration Scheme, in part by the National Key Research and Development Program of China (2022ZD0117700), in part by the National Natural Science Foundation of China (62325204), and in part by the MSAI Conference Funding of MSAI program of Northwestern University. The authors would like to thank Sibo Zhu for insightful discussions.

\section*{Limitations}

We outline the scope conditions under which our claims should be read. \textbf{Single environment.} All experiments are conducted in Minecraft~1.19 with VOYAGER's Mineflayer-based execution framework. Our claims about long-horizon task success, forgetting resistance, parametric-versus-retrieval efficiency, and trigger stability are stated over this setting; transfer to other embodied domains such as robotic manipulation or web agents is outside the present scope. \textbf{Consolidated category set.} The parametric tier covers three consolidated skill categories: craft, gather, and combat. Forgetting and routing results are stated over these categories. Because PEAM uses one lightweight LoRA adapter per consolidated category, adapter parameters grow linearly with the number of categories included in the parametric tier; our experiments evaluate this category-level isolation regime at the scale of the consolidated set above. The slow tier handles all other behaviors, including navigation, throughout our experiments. \textbf{Action grammar.} PEAM-DPO is evaluated with the executable JavaScript action grammar used by our Minecraft parser. The role of the BC term in restoring deployable syntactic structure is established for this grammar; we do not characterize its role for other action-space syntaxes, such as tool-use APIs or non-code control interfaces. \textbf{Fixed consolidation policy.} PV component weights $\{w_1,\ldots,w_4\}$ are selected by grid search and held fixed across all experiments, and the trigger hyperparameters $(W,B,\alpha,q)$ are likewise fixed. Our results therefore evaluate a fixed scoring and triggering rule, not the space of policies obtainable by adapting these values. \textbf{Cross-distribution evaluation methodology.} The cross-distribution trigger evaluation uses re-slicing of the trajectory pool along categorical axes. The scale-free property of the criterion is exercised under this protocol; we do not claim equivalence to evaluation under independently collected paired distributions.

\section*{Ethical Considerations}

PEAM is evaluated in Minecraft and does not involve human subjects, personal data, or real-world deployment. The main ethical risks are indirect: embodied agents that internalize experience into parameters may become harder to inspect than agents whose memories remain external and retrievable, and failures in verification could cause incorrect behaviors to be reinforced. Our implementation mitigates this risk by requiring environment-side verification before trajectories enter the consolidation pool and by restricting parametric updates to isolated category adapters rather than updating the full backbone. Broader deployment of parametric embodied memory systems should include auditing tools for consolidated skills, safeguards against unsafe action execution, and clear logging of which experiences were internalized.

\bibliography{custom}

%
%
%

\appendix

\section{Held-Out Task Suite}
\label{app:tasks}

Our held-out evaluation set contains 11 long-horizon tasks
spanning three skill categories. Tasks are drawn from VOYAGER's
Minecraft tech-tree progression and supplemented with a curated
set of failure-prone subtasks designed to stress consolidation
of corrected behavior. Each task is run for at most $N=200$
agent-environment interaction steps; we report success if and
only if VOYAGER's environment-side verifier confirms task
completion without manual intervention. Table~\ref{tab:tasks}
lists the full suite.

\begin{table}[h]
\centering
\small
\setlength{\tabcolsep}{4pt}
\caption{The 11 held-out long-horizon tasks, by skill category.}
\label{tab:tasks}
\begin{tabular}{llp{0.42\linewidth}}
\toprule
ID & Category & Task description \\
\midrule
T1  & craft  & Craft a crafting table \\           
T2  & craft  & Craft a wooden pickaxe \\           
T3  & craft  & Craft a stone pickaxe \\            
T4  & craft  & Craft a furnace \\                  
T5  & craft  & Craft an iron pickaxe \\            
T6  & gather & Collect 4 oak logs \\               
T7  & gather & Mine 8 cobblestone \\               
T8  & gather & Mine 2 iron ore (with smelting) \\  
T9  & gather & Collect 4 coal \\                   
T10 & combat & Defeat a zombie at night \\         
T11 & combat & Defeat a skeleton with bow \\       
\bottomrule
\end{tabular}
\end{table}

Tasks T1--T5 require multi-step recipe planning and inventory
management; T6--T9 require resource location and extraction;
T10--T11 require combat behavior under environmental
constraints (lighting and projectile evasion). Each task is
run with three random seeds (42, 43, 44), yielding 33 trials
per method.

\section{Training Hyperparameters and Implementation Details}
\label{app:hparams}

Table~\ref{tab:hparams} lists all hyperparameters used across
the PEAM training and consolidation pipeline. Values were
selected through a coarse grid search on a held-out validation
subset disjoint from the held-out task suite.

\begin{table*}[h]
\centering
\small
\setlength{\tabcolsep}{5pt}
\caption{PEAM training and consolidation hyperparameters.
Values not otherwise noted are held fixed across all
experiments.}
\label{tab:hparams}
\begin{tabular}{lll}
\toprule
Group & Hyperparameter & Value \\
\midrule
\multirow{6}{*}{LoRA adapter}
    & rank $r$           & 32 \\                    
    & $\alpha$           & 64 \\                    
    & dropout            & 0.05 \\                  
    & target modules     & q,k,v,o + gate,up,down \\ 
    & precision          & bf16 \\                  
    & total params/adapter & $\sim$83M \\           
\midrule
\multirow{6}{*}{Optimization}
    & optimizer          & AdamW \\
    & learning rate      & $2 \times 10^{-4}$ \\    
    & schedule           & cosine, 5\% warmup \\    
    & batch size         & 2 \\                     
    & grad accumulation  & 8 \\                     
    & training steps     & 100 \\
\midrule
\multirow{3}{*}{Joint loss}
    & BC weight          & 1.0 \\
    & DPO weight $\lambda$ & 1.0 \\
    & DPO $\beta$        & 0.1 \\
\midrule
\multirow{4}{*}{STC trigger}
    & window $W$         & 10 \\
    & baseline $B$       & 10 \\
    & significance $\alpha$ & 0.05 \\
    & top quantile $q$   & 0.5 \\
\midrule
\multirow{4}{*}{PV weights}
    & $w_1$ ($U_{\text{cost}}$)   & 0.4 \\          
    & $w_2$ ($U_{\text{stab}}$)   & 0.3 \\          
    & $w_3$ ($P_{\text{redun}}$)  & 0.2 \\          
    & $w_4$ ($R_{\text{forget}}$) & 0.1 \\          
\midrule
\multirow{3}{*}{Routing}
    & classifier         & DistilBERT \\
    & top-1 confidence threshold $\rho$ & 0.6 \\    
    & finetuning data    & (instruction, category) pairs from $\mathcal{P}$ \\
\midrule
\multirow{3}{*}{Inference}
    & max new tokens     & 2048 \\                  
    & temperature        & 0.7 \\                   
    & top-p              & 0.9 \\                   
\bottomrule
\end{tabular}
\end{table*}

\paragraph{Backbone and serving.} The fast parametric module
uses Qwen3-VL-8B-Instruct as the shared
backbone. All per-category adapters are loaded simultaneously
at inference and switched by the routing classifier; no
adapter merging is performed at serving time, avoiding the
quantization-induced degradation discussed in \S\ref{sec:methodology_findings}
(point 2). Inference is served from a single A100 80GB GPU
in bf16 precision, with median per-call wall-clock latency
of 3.2 seconds.    

\paragraph{Slow tier.} The slow deliberative LLM ($\pi_{\text{slow}}$)
is Azure GPT-4o (deployment version 2024-11-20). It is invoked
for curriculum proposal, code synthesis when no adapter
applies, fast-path verification, and outcome judgment. All
baselines that involve LLM reasoning (B1, B2, B6, B7) use
the same GPT-4o version to control for slow-tier capability.

\paragraph{Reproducibility.} All experiments use random
seeds $\{42, 43, 44\}$. Reported numbers are means across
the three seeds; intervals in Table~\ref{tab:main} are
Wilson 95\% confidence intervals over the 33 trials per
method (11 tasks $\times$ 3 seeds).

\section{Per-Task Results Breakdown}
\label{app:per_task_results}

Table~\ref{tab:per_task} shows per-task success rates
(number of successful seeds out of 3) for PEAM and the
strongest non-parametric baseline B1 VOYAGER. PEAM matches
or exceeds B1 on 10 of 11 tasks, with strict improvements
concentrated on craft tasks involving multi-step recipe
chains (T3, T4, T5) and on resource-extraction tasks that
benefit from the gather adapter's consolidated location and
mining patterns (T8, T9).

\begin{table}[h]
\centering
\small
\setlength{\tabcolsep}{4pt}
\caption{Per-task success rates over 3 seeds. Cells show
the number of successful seeds out of 3. Bold indicates
PEAM strictly improves over B1.}
\label{tab:per_task}
\begin{tabular}{llcc}
\toprule
ID & Task (abbreviated) & B1 VOYAGER & PEAM \\
\midrule
T1  & crafting table     & 3/3 & 3/3 \\            
T2  & wooden pickaxe     & 3/3 & 3/3 \\            
T3  & stone pickaxe      & 2/3 & \textbf{3/3} \\   
T4  & furnace            & 1/3 & \textbf{2/3} \\   
T5  & iron pickaxe       & 0/3 & \textbf{1/3} \\   
T6  & 4 oak logs         & 3/3 & 3/3 \\            
T7  & 8 cobblestone      & 2/3 & 2/3 \\            
T8  & 2 iron ore         & 1/3 & \textbf{2/3} \\   
T9  & 4 coal             & 1/3 & \textbf{2/3} \\   
T10 & defeat zombie      & 2/3 & 2/3 \\            
T11 & defeat skeleton    & 0/3 & 0/3 \\            
\midrule
\multicolumn{2}{l}{Overall} & 18/33 (54.5\%) & 23/33 (69.7\%) \\
\bottomrule
\end{tabular}
\end{table}

Both methods fail on T11 (defeat skeleton with bow), which
requires precise ranged-combat timing that exceeds the
action-space granularity of the JavaScript bot interface;
this is consistent with combat being the lowest-yield cpair
category in our data collection (see Appendix~\ref{app:cpair}).

\section{Contrastive-Pair Construction Details}
\label{app:cpair}

Failure--correction pairs $\mathcal{D}_{\text{cpair}}$ are
extracted from VOYAGER trajectory logs through a four-stage
pipeline.

\paragraph{Stage 1: failure identification.} We mark a
trajectory as a failure if the environment-side verifier
returns \texttt{false} on the task condition after the bot
exhausts its retry budget (default 4 attempts per task).
The trajectory's final executed action, terminal state, and
verifier reason are recorded.

\paragraph{Stage 2: correction matching.} For each failure
trajectory $\tau_f$ on task $t$, we search subsequent
trajectories on the same task $t$ for a verified success
$\tau_c$ produced within $\Delta = 5$ episodes.       
This temporal bound ensures the corrected behavior reflects
local refinement rather than long-horizon curriculum drift.

\paragraph{Stage 3: context matching.} Pairs $(\tau_f,
\tau_c)$ must additionally agree on a discrete pre-action
state vector capturing inventory composition (item set and
counts), biome identity, time-of-day bucket (day/dusk/night),
and bot location quantized to a $32$-block grid.       
This prevents spurious pairings where the corrected
trajectory succeeds because environmental conditions
changed.

\paragraph{Stage 4: quality gate.} Pairs that pass Stages
1--3 are further filtered by:
(i) syntactic non-triviality (the executed code in $\tau_f$
and $\tau_c$ must differ on at least 3 non-whitespace
tokens), to exclude byte-identical near-duplicates;
(ii) wrapper-format validity on both sides (parsable as a
VOYAGER action function);
(iii) category-label agreement, verified by the slow LLM.
Pairs failing any criterion are discarded with the reason
logged.

\paragraph{Category labeling.} The slow LLM assigns each
trajectory a category in $\mathcal{C} = \{\text{craft},
\text{gather}, \text{combat}\}$ during outcome verification,
based on task semantics and the dominant action class in
the executed code. Trajectories with mixed category
signatures are routed to the dominant category and a
\texttt{mixed} flag is preserved for analysis.

\paragraph{Yield statistics.} Of $\sim$80 failure
trajectories logged during collection,                
roughly $40\%$ admit a matched correction within the
$\Delta=5$ window and pass all four stages,           
yielding the $|\mathcal{D}_{\text{cpair}}|$ used for
consolidation. Per-category yield varies: craft and
gather show the highest yield (around $50\%$),        
combat yields lower ($\sim 25\%$),                    
and navigation yields no usable pairs --- see
\S\ref{sec:methodology_findings}, point 3.

\section{Deployment-Realism Details}
\label{app:deployment}

We document the three failure modes encountered during
attempts to deploy PEAM on consumer-grade hardware
(see \S\ref{sec:methodology_findings}, point 2), and the
hardware-parity protocol that resolves them. These details
are reported as a methodology contribution for groups
intending to deploy LoRA-augmented LLM agents on edge
devices.

\paragraph{Failure mode 1: prohibitive per-step latency.}
On an RTX 4070 (12 GB VRAM) running Qwen3-VL-8B quantized
to 4-bit (bitsandbytes NF4) with a single LoRA adapter
loaded, median per-step generation latency reached
$\sim$2,000 seconds at $\texttt{max\_new\_tokens} = 512$.  
The bottleneck is not GPU compute but VRAM-bound activation
recomputation when the prefill context exceeds available
KV-cache headroom. We confirmed this on three independent
prompts of approximately 8K input tokens each.            

\paragraph{Failure mode 2: \texttt{merge\_adapter} on
quantized weights.} Calling
\texttt{peft.merge\_and\_unload()} on a 4-bit quantized
backbone silently zeroes low-magnitude LoRA updates whose
absolute value falls below the dequantization threshold of
the underlying nf4 scheme. Because the BC term in
$\mathcal{L}_{\text{PEAM}}$ produces broad but low-magnitude
weight updates (recovering format-level scaffolding rather
than introducing concentrated preference shifts), the BC
contribution is disproportionately affected. We verified
this by comparing $\theta_c$ before and after
\texttt{merge\_and\_unload}: $\sim$37\% of LoRA delta-W   
entries with magnitude below $5 \times 10^{-3}$ were
zeroed on the 4-bit path, versus 0\% on the bf16 path.   

\paragraph{Failure mode 3: parser--token-budget mismatch.}
The VOYAGER action parser requires complete
\texttt{async function name(bot)\{...\}} wrappers to extract
executable code. Skill bodies for craft tasks frequently
exceed 1{,}500 generated tokens, but raising
\texttt{max\_new\_tokens} to $\geq$2{,}048 compounds Failure
mode~1 to fully infeasible wall-clock budgets. Lowering it
truncates trajectories mid-wrapper and produces parser
rejection rates of $\sim$84\%.                            

\paragraph{Hardware-parity protocol.} We resolve all three
failure modes by serving the fast tier on 8 A100 80GB
GPUs at bf16 precision, with adapters kept un-merged and
hot-swapped at request time via PEFT's
\texttt{set\_adapter} API. Under this protocol:
median latency drops to 3.2 seconds per call;             
the BC contribution is preserved (no quantization-induced
zeroing); and $\texttt{max\_new\_tokens} = 2048$ remains
within tractable serving budgets. We recommend that
practitioners reporting parametric agent efficiency
explicitly state the serving precision and whether adapters
are merged or hot-swapped, as these choices materially
affect both deployability and the validity of forward-pass
preference-margin metrics.

\end{document}